\documentclass[manuscript,screen,nonacm,acmtog]{acmart}

\newcommand{\calL}{{\mathcal{L}}}
\newcommand{\calX}{{\mathcal{X}}}

\newcommand{\calD}{{\mathcal{D}}}

\newcommand{\calY}{{\mathcal{Y}}}

\newcommand{\one}{\boldsymbol{1}}

\newcommand{\p}{\bm{p}}
\newcommand{\q}{\bm{q}}

\newcommand{\R}{\mathbb{R}}

\newcommand{\x}{\bm{x}}

\newcommand{\E}{\mathbb{E}}

\newcommand{\bu}{\bm{u}}

\newcommand{\smx}{\sigma}

\newcommand{\eat}[1]{}
\newcommand{\hush}[1]{}

\usepackage{microtype}
\usepackage{graphicx}
\usepackage{subfigure}
\usepackage{booktabs} 

\usepackage{hyperref}

\newcommand{\Looper}{Platform}

\usepackage{amsmath}
\usepackage{mathtools}
\usepackage{amsthm}
\usepackage{bm}

\usepackage{multirow}

\usepackage[utf8]{inputenc} 
\usepackage[T1]{fontenc}    
\usepackage{hyperref}       
\usepackage{url}            
\usepackage{booktabs}       
\usepackage{amsfonts}       
\usepackage{nicefrac}       
\usepackage{microtype}      
\usepackage{bbm}
\usepackage[capitalize,noabbrev]{cleveref}

\theoremstyle{plain}

\theoremstyle{definition}

\theoremstyle{remark}

\begin{document}

\title{Practical Knowledge Distillation: Using DNNs to Beat DNNs}

\author{Chung-Wei Lee}
\email{leechung@usc.edu}
\affiliation{
  \institution{USC}
  \city{Los Angeles}
  \state{CA}
  \country{USA}
  \postcode{90098}
}

\author{Pavlos Athanasios Apostolopulos}
\email{pavlosapost@meta.com}
\author{Igor L. Markov}
\email{imarkov@meta.com}
\affiliation{
  \institution{Meta}
  \city{Menlo Park}
  \state{CA}
  \country{USA}
  \postcode{94025}
 }








\begin{abstract}
For tabular data sets, we explore data and model distillation, as well as data denoising.
These techniques improve both gradient-boosting models and a specialized DNN architecture. While gradient boosting is known to outperform DNNs on tabular data, we close the gap for datasets with 100K+ rows and give DNNs an advantage on small data sets. We extend these results with input-data distillation and optimized ensembling to help DNN performance match or exceed that of gradient boosting. As a theoretical justification of our practical method, we prove its equivalence to classical cross-entropy knowledge distillation. We also qualitatively explain the superiority of DNN ensembles over XGBoost
on small data sets. For an industry end-to-end real-time ML platform with 4M production inferences per second, we develop a model-training workflow based on data sampling that distills ensembles of models into a single gradient-boosting model favored for high-performance real-time inference,
without performance loss. Empirical evaluation shows that the proposed combination of methods consistently improves model accuracy over prior best models across several production applications deployed worldwide.
\end{abstract}



\maketitle



\section{Introduction}

As machine learning is applied to more applications, higher-quality models require improving data quality,
increasing data volume, using more sophisticated model types, optimizing hyperparameters, and training more effectively. These paths are backed by theory, but practical contexts are not always in line with assumptions in formal analyses, while 
real-world considerations may not favor the models
preferred by theory. Such disconnects can be illustrated by the significance of resource consumption at inference in real-time ML platforms that favors simpler models. In terms of model quality, gradient boosting tends to outperform Deep Learning models on {\em tabular data}, for reasons that have been carefully analysed only recently \cite{grinsztajn2022tree,shwartz2022tabular,Gorishny2021revisiting}. 
Moreover, combining multiple methods tends to give better results in practice \cite{ganaie2021ensemble}. Hence, our work is motivated by practical considerations and uses tabular data from \Looper, a high-performance end-to-end real-time ML platform~[removed for blind review]
but also offers theoretical insights that match and explain empirical results. Hundreds of production use cases running on this platform cumulatively perform millions of inferences per second. Hence, both ML quality and the efficiency of inference are critical. Real-time inference is valuable to many applications on this platform, but limits the types of models during inference. In particular, gradient boosting models are often favored for their small memory footprint and fast inference, whereas model ensembles are not supported easily. While\eat{the broad adoption of machine learning and} vibrant ML research introduces new model architectures daily, an industry ML platform cannot realistically support high-performance inference for all new model types that show promising results during training. Our work shows that model distillation helps not only to improve an individual models' performance, but also to leverage new model types as teachers to train simpler student models fully supported by the platform. In many production use cases, labels 
capture human behavior or the environment, and therefore suffer noise, inconsistencies and other irregularities. Generic data denoising \cite{song2022learning, 9784878, chen2021learning} and data distillation \cite{medvedev2021new, larasati2022review} methods may be applied before model training  to improve ML models. But such methods must be automated when production ML models are retrained daily to handle nonstationary data distributions~\cite{markov2022looper}.

To address relevant ML applications, we explore recent advancements
in ML research: ($i$) data and model distillation, ($ii$) denoising and uncertainty mitigation, and ($iii$) deep tabular learning.

\begin{figure*}[t]
\includegraphics[width=0.8\paperwidth,trim={4mm 6cm 4mm 0},clip]{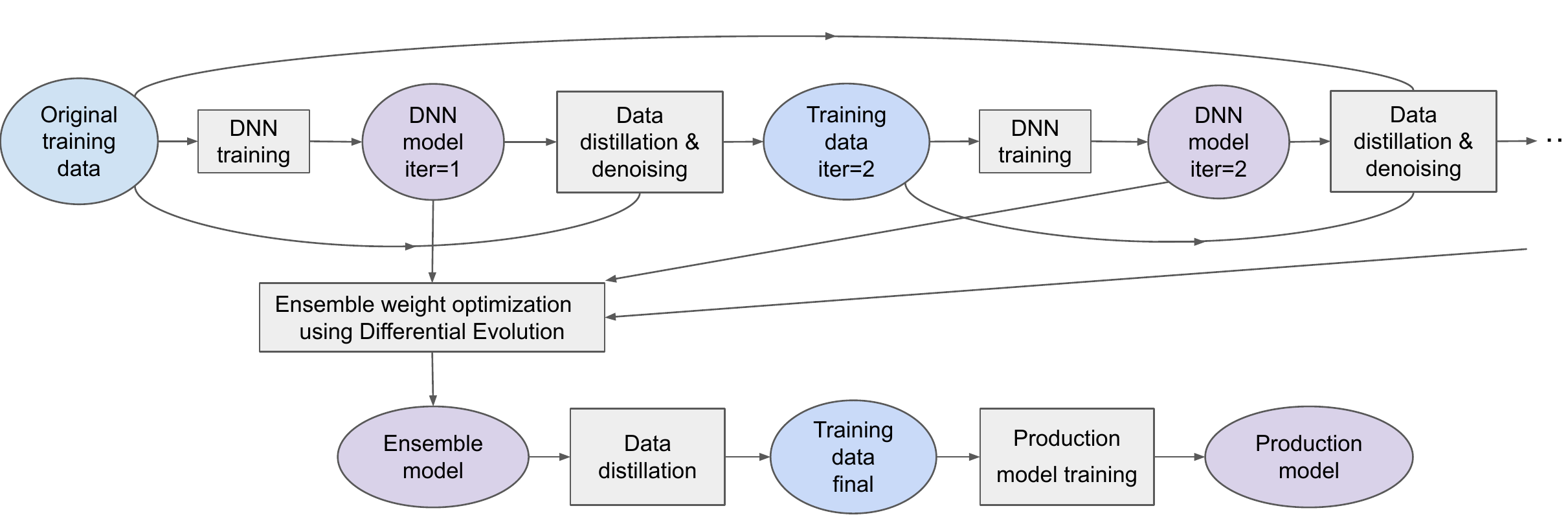}
\caption{
\label{fig:workflow}
High-level workflow developed in this paper.
Original training data assumes discrete labels.
Distilled data replaces labels with model scores,
which include {\em dark knowledge}. Denoising removes
data rows where model scores differ too much from input labels
or scores. Iterations of model training and data distillation
repeat, while training models of the same architecture and capacity (self-distillation).
Each new model is trained using the last iteration of distilled and 
denoised training data as well as the original training data. Model quality improvement is not required for each new model. An ensemble model combines all intermediate models and, possibly, models of additional types trained on original data. A weight-optimized ensemble model is distilled again into a simpler production model.
}
\end{figure*}

\noindent
{\bf Knowledge distillation} \cite{hinton2015distilling} was originally proposed to transfer knowledge from a “teacher” model with a larger capacity to a “student” model with a smaller capacity, to achieve {\em model compression}. When training a student model, an extra term in the loss function encourages the student model to mimic the teacher model. 
Surprisingly, the paper on Born-Again Neural Networks \cite{furlanello2018born} shows that when the student uses the same architecture and capacity as the teacher model, the student model may outperform the teacher using {\em dark knowledge} (a term introduced by G.Hinton) extracted from the teacher model. Moreover, the performance is further enhanced by averaging the predictions from multiple sequentially-trained student models (where the student becomes a teacher) and by using knowledge distillation (i.e., building an ensemble model). This generic way to boost an ML model has only been evaluated for DNN models so far.  

\noindent
{\bf Uncertainty and noise} are frequently observed in practical ML applications in several forms. An important distinction is between {\em epistemic}  and {\em aleatoric} uncertainty \cite{hullermeier2021aleatoric}. Epistemic uncertainty is associated with insufficient information carried by a given data sample (drawn randomly or not) and can be reduced by adding more data --- training ML models on larger datasets often improves results. In contrast, aleatoric uncertainty is associated with individual data rows and cannot be mitigated by adding more data. The practical model-training workflow deveveloped and evaluated in this paper addresses aleatoric uncertainty (including label noise) by dropping a small fraction of carefully selected data rows. Additionally, we demonstrate that ensembles of distilled DNNs hold significant advantage over gradient boosting models on small training datasets, where gradient boosting overfits but DNN ensembles generalize well and thus address epistemic uncertainty.

\noindent
{\bf Tabular data and deep learning}.
Much of ML literature on label denoising and knowledge distillation deals with image data and DNN-based model architectures. However, industrial applications like ours deal with tabular data by building better-performing gradient boosting models, such as XGBoost \cite{chen2016xgboost}. In our work, we consider both XGBoost and TabNet \cite{arik2021tabnet}, a recent DNN architecture for tabular data. TabNet employs the {\em attention} mechanism for soft feature selection, enabling interpretability, and has shown performance competitive to XGBoost on public tabular datasets. Given that XGBoost dominates DNNs for tabular data in comparisons on open datasets \cite{grinsztajn2022tree,shwartz2022tabular,Gorishny2021revisiting}
and in industry ML platforms \cite{markov2022looper},
one of our goals is to explore if TabNet (or more sophisticated models built from it) can compete with XGBoost on tabular datasets. 

Our emphasis on tabular data has significant impact not only 
on the choice of ML model type, but also on the applicability of denoising methods. In particular, denoising methods often exploit 
correlated features (such as image pixels) and redundant information (from nearby video frames, from different parts of an image sufficient for classification decisions, etc). However, features in tabular data are
typically unrelated and may be on different scales, while redundancies are uncommon. Less obviously, the diversity of feature/column types in tabular data undermines the notions of {\em distance} and {\em similarity} that are natural to multimedia data. In turn, this invalidates denoising methods that
rely on K nearest neighbors, similarity clustering and other distance-based techniques (that assume that similar inputs must lead to similar correct outputs).

\noindent
{\bf In this work},
we combine denoising with data and model distillation in the context of deep tabular learning. We observe that TabNet underperforms XGBoost on larger datasets (100K rows). This is counterintuitive as DNNs are assumed to perform well on large datasets, but recent literature made similar observations and explained them \cite{shwartz2022tabular,grinsztajn2022tree,Gorishny2021revisiting}. With an eye on easy-to-implement optimizations of TabNet-based models, we assemble a series of practical improvements (Figure \ref{fig:workflow}).
Empirical validation indicates that resulting TabNet ensembles are competitive with XGBoost and in some cases superior to XGBoost. Ensembling them with XGBoost produces even better models. However, supporting such mixed ensemble for high-performance real-time inference runs into platform limitations. We then use our distillation via weighted datasets to distill across models of different types: we distill our best XGBoost + TabNet ensemble into a single XGBoost model with no loss of performance. Implementation is easy and helps us outperform ML models previously deployed on the ML platform\hush{\cite{markov2022looper}} for all tested use cases (Table \ref{tab:final}). More generally, in an end-to-end ML platform,
this simple shortcut can extend the benefits of new ML model types with promising training results
toward high-performance inference without impacting the complexity of inference infrastructure.

\section{Related Work}

\noindent
{\bf Knowledge distillation.}
Large deep neural networks \cite{https://doi.org/10.48550/arxiv.1512.03385, devlin2018bert} have led to impressive feats in real-world applications with large-scale data \cite{you2019large}. However, their deployment in industry-scale applications for real-time inference, poses challenges due to computational complexity, storage requirements and inference latency. 
In response, knowledge distillation (KD) \cite{hinton2015distilling} was introduced as a model-compression technique that distills the knowledge from a {\em larger} neural network to a {\em smaller} network \cite{urban2016deep, sun2019patient, panchapagesan2021efficient}. Despite KD's great success in natural language processing, computer vision, etc \cite{alkhulaifi2021knowledge, gou2021knowledge}, its theoretical and empirical understanding remains limited \cite{cheng2020explaining, phuong2019towards}. Also, while KD trains student models in supervision of a teacher model, researchers developed several promising domain-specific learning schemes \cite{gou2021knowledge} that select teacher's components and determine how they are used in training: {\em response-based knowledge} \cite{meng2019conditional}, {\em feature-based knowledge} \cite{wang2020exclusivity}, and {\em relation-based knowledge} \cite{chen2020learning}. Other practical KD techniques include {\em adversarial distillation} \cite{mirzadeh2020improved}, {\em multi-teacher distillation} \cite{yang2020model}, {\em cross-modal distillation} \cite{zhao2020knowledge}. Applications of these algorithms often rely on heavily-tuned implementations \cite{zhang2020knowledge} whereas many practical uses encourage full automation to support regular model retraining to follow nonstationary data. Our work largely draws on {\em self-distillation}, where the architecture and capacity of the network does not
change during training.

\noindent
{\bf Training on noisy data}, is widely recognized as a practical circumstance that may limit success of Machine Learning (ML) \cite{nigam2020impact}. In the context of DNNs, the problem is covered well in the literature, due to both the popularity of DNNs and DNN's propensity to overfit~\cite{krause2016unreasonable}. For instance, the work in \cite{zhang2021understanding} shows that sufficiently large DNNs can easily fit an entire training dataset with any ratio of corrupted labels. The noisy-label problem can be viewed from several perspectives \cite{learning-from-noise-dnn}. The {\em feature-class dependency} checks if noise levels dependent on the labels alone or also on the features. The {\em open-close dichotomy} arises when the set of possible labels is large (open set) and a correct label may not even be observed, whereas the {\em close-set} alternative assumes correct labels to be among observed labels.
In any case, training with noise is often related to the {\em memorization effect} of deep learning \cite{arpit2017closer}, and proposed mitigation techniques usually seek to downweigh or skip errouneous labels. Solutions have been inspired by the freedom to select DNN model architectures \cite{co-teaching-noise}, loss-based regularization conditions \cite{patrini2017making}, semi-supervised learning \cite{semi-supervised-noise} and meta-learning \cite{shu2019meta}. Common limitations include limited scalability and and specificity to DNN architectures \cite{9474624,industrial-label}. Techniques to deal with noise usually rely on the assumption that similar inputs (images with similar pixels, words or sentences with similar embedding vectors) must produce the same model output. However, {\em tabular data} often lack a well-defined notion of distance (e.g., different timestamps added to a data row may be superflouous in some cases and important in other cases; also, combining floating-point features with categorical features to define a distance is ambiguous). Therefore, many existing techniques for training with noise are not helpful with tabular data.

\noindent
{\bf Ensemble learning}, such as averaging the outputs of several models, is common in ML practice and especially in ML contests \cite{ensemble-learning-survey}. An {\em ensemble} of ML models combines predictions of multiple constituent models, possibly of different types, to diminish overfitting. 
This tends to offer greater benefits for DNNs
\cite{Yang_2022}, likely because DNNs are initialized with random weights,
leading to diverse DNNs with comparable ML performance. While simple averaging of models often shows promising results, linear and non-linear functions can aggregate models using additional parameters that can be optimized~\cite{ensemble-optimization, ensemble-optimization-pricing}. For example, averaging can be extended by adding a handful of tunable weights to optimize, e.g., the performance of the resulting classifier. Traditional optimization methods would be difficult to apply here because the dependence of the end metric on the parameters is not captured by an explicit formula. Therefore, one resorts to gradient-free black-box methods, such as Differential Evolution \cite{AHMAD20223831} that maintains a population of configurations and creates new configurations using a randomized crossover operator.

\hush{The rapid advancement of deep learning has led to impressive feats in applications such as computer vision and natural language processing \cite{https://doi.org/10.48550/arxiv.2203.02155}. Often, the training of very deep models \cite{https://doi.org/10.48550/arxiv.1512.03385, devlin2018bert} with thousands of layers, is necessary given the complexity and volume of data \cite{you2019large}. Although techniques that enable the training of such deep models exist \cite{he2020resnet, ioffe2015batch}, their huge computational complexity, massive storage requirements, and inference latency make it challenging to deploy such models in production for real-time inference in industry-scale applications.}

\section{Problem Setting and Motivation}
\label{sec:motivation}
We focus on binary classification with the ROC AUC objective, for simplicity. Consider $N$ training samples $\{(x_i,y_i)\}$, where $x_i \in R^d$ and $y_i \in\{0,1\}$. A machine learning model implements a function $f: \mathbb{R}^n \to [0,1]$ that maps each $x_i$ to $\text{Prob}[y_i = 1]$.

\noindent
A model $f_0$ trained on the original dataset $\{(x_i,y_i)\}$ is termed a {\em teacher} model. The key idea of model distillation is then to transfer the knowledge of the {\em teacher} $f_o$ to a {\em student} model $f_1$. 
The latter is traditionally done
by training with a cross-entropy term between the student’s and the teacher’s predictions:
$$\sum_{i=1}^N - f_0(x_i) \log(f_1(x_i)) -  f_0(1-x_i) \log(f_1(1-x_i)).$$
Adding this term to the loss function encourages the student model to mimic the teacher’s output. Needless to say, not every ML model can accommodate this approach. Additionally, our industry
setting encourages black-box components with clean interfaces. To this end, we explore a different approach to transfer knowledge from the teacher model --- we train $f_1$ on a weighted dataset, where each $(x_i,0)$ has a weight $1-f_0(x_i)$ and each $(x_i,1)$ has a weight $f_0(x_0)$.
We term this approach \emph{input-data distillation}.
Intuitively, the weighted dataset summarizes the teacher model’s belief on the training data. In practice, the model-training pipeline can interpret the weights as {\em row-sampling probabilities for the data loader} (when forming a batch), which enables this approach with any eligible ML model that might not even support training with weighted datasets.


\subsection{Motivational Experiments on image datasets}

We first replicate the results of knowledge distillation with Born-again networks \cite{furlanello2018born} using authors’ code, on open {\em image} datasets MNIST and CIFAR10 (60K images) using ResNet50, we also add XGBoost. In Table \ref{tab:CIFAR_MNIST}, the figure of merit is accuracy. We observe that distillation does not necessarily improve individual student models, but their ensembles demonstrate consistent improvement. In each column, the best results (highlighted) are attained by ensembles. Thus,
 knowledge distillation with ensembles promises to improve initial models, but the DNN model (ResNet50) has more to gain than XGBoost (we return to this
trend later).

\begin{table}[!ht]
\centering\caption{XGBoost and ResNet performance on CIFAR10 and MNIST with knowledge distillation; generations and ensembles are indicated in rows. The best-performing row is highligted.}
\label{tab:CIFAR_MNIST}
\begin{tabular}{|c|l|l|l|l|}
\hline
Dataset           & CIFAR10                                 & CIFAR10                                 & MNIST                                   & MNIST                                   \\
Model             & XGBoost                                 & ResNet50                                & XGBoost                                 & ResNet50                                \\ \hline
Gen 0 ($f_0$) & 0.5440                                  & 0.8339                                  & 0.9735                                  & 0.9901                                  \\ \hline
Gen 1 ($f_1$) & 0.5356                                  & 0.8257                                  & 0.9745                                  & 0.9864                                  \\ \hline
Gen 2 ($f_2$) & 0.5284                                  & 0.8358                                  & 0.9734                                  & 0.9892                                  \\ \hline
Ens 0+1      & 0.5499                                  & 0.8593                                  & 0.9749                                  & 0.9913                                  \\ \hline
Ens 0+1+2    & \textbf{0.5517} & \textbf{0.8703} & \textbf{0.9752} & \textbf{0.9922} \\ \hline
\end{tabular}
\end{table}

\subsection{Motivational experiments on tabular datasets}

In this paper, we focus on {\em tabular} data sets deployed in production use cases on the high-performance end-to-end real-time ML platform.
Unlike in image data,
columns of tabular data are often at different scales, may be of different types, may carry different amounts of noise, and do not correlate like nearby image pixels do \cite{shwartz2022tabular,grinsztajn2022tree,Gorishny2021revisiting}. Shift-invariance 
(a letter ``O'' can be moved by several pixels up or down)
rarely arises in tabular data, which negates the advantages of ConvNets.

Given that we study binary classification, the figure of merit is ROC AUC. A collection of \Looper data sets is described in Table \ref{tab:meta data}. These data sets represent applications deployed ($i$) in a major social network, ($ii$) on an internal corporate system, ($iii$) in a popular augmented-reality system. They are used for final evaluation in Table \ref{tab:final}.
For evaluating individual improvements, we found \texttt{Dataset B} representative and complement it with other datasets on occasion (the text refers to table in the Appendix). We start by replicating past work and testing simple ideas, as groundwork for more advanced techniques in Section \ref{sec:advanced}. For all experiments with DNN models, we use TabNet \cite{arik2021tabnet} (the implementation posted by the authors for public use) and compare it with XGBoost \cite{chen2016xgboost}.

\noindent 
{\bf TabNet gains more from ensembling than XGBoost does, but still lags behind.}
We first evaluated plain TabNet and XGBoost “out of the box” without any hyperparameter tuning. Per Table \ref{tab:ensemblesB100K}, both benefited from knowledge distillation, but the gain for TabNet was greater than for XGBoost. XGBoost still outperforms TabNet.
We observed similar results on other datasets and with the {\em patience} hyperparameter for TabNet increased by 3$\times$ (by default, TabNet stops training after 10 consecutive nonimproving epochs). 

\begin{table}[!ht]
\centering\caption{XGBoost and TabNet ensembles on Dataset B (100K sampled rows)}
\label{tab:ensemblesB100K}
\begin{tabular}{|c|l|l|l|l|}
\hline
Gen & \begin{tabular}[c]{@{}l@{}}TabNet\\ AUC (n)\end{tabular} & \begin{tabular}[c]{@{}l@{}}TabNet Ens \\ AUC (1+..+n)\end{tabular} & \begin{tabular}[c]{@{}l@{}}XGBoost\\ AUC (n)\end{tabular} & \begin{tabular}[c]{@{}l@{}}XGBoost\\ Ensemble\\ AUC (1+..+n)\end{tabular} \\ \hline
0   & 0.60982                                                  &                                                                         & 0.67722                                                   &                                                                           \\ \hline
1   & 0.61859                                                  & 0.61995                                                                 & 0.70136                                                   & 0.70052                                                                   \\ \hline
2   & 0.58359                                                  & 0.61519                                                                 & 0.72189                                                   & 0.71317                                                                   \\ \hline
3   & 0.57927                                                  & 0.61867                                                                 & 0.72304                                                   & 0.71944                                                                   \\ \hline
4   & 0.57336                                                  & 0.6233                                                                  & 0.72406                                                   & 0.72273                                                                   \\ \hline
5   & 0.57793                                                  & \textbf{0.62547}                                & 0.72369                                                   & \textbf{0.72418}                                  \\ \hline
\end{tabular}
\end{table}

\noindent
{\bf Removing {\em noisy} labels improves performance} --- a simple, but powerful idea. When trainig with knowledge distillation, we remove data rows whose predictions are too far from the teacher model’s predictions, that is, we remove data $i$ if  $\Delta=|f(x_i)-y_i| \ge$ threshold. A threshold equal to 1 keeps all rows. Per Table \ref{tab:denoisingB}, 
threshold 0.99 leads to good performance and consistently improves results. 

\begin{table*}[!ht]
\vspace{5mm}
\centering\caption{Data denoising during TabNet distillation using different thresholds on Dataset B}
\label{tab:denoisingB}

\begin{tabular}{|c|ll|ll|ll|ll|ll|}
\hline
 
\multicolumn{1}{|l|}{} & \multicolumn{2}{c|}{threshold = 1}                                            & \multicolumn{2}{c|}{threshold=0.999}                                                    & \multicolumn{2}{c|}{threshold=0.99}                                                           & \multicolumn{2}{c|}{threshold=0.9}                                                      & \multicolumn{2}{c|}{threshold=0.5}                                                      \\ \hline
 
Gen                                            & \multicolumn{1}{c|}{Individual} & \multicolumn{1}{c|}{Ensemble} & \multicolumn{1}{c|}{Individual} & \multicolumn{1}{c|}{Ensemble} & \multicolumn{1}{c|}{Individual}       & \multicolumn{1}{c|}{Ensemble} & \multicolumn{1}{c|}{Individual} & \multicolumn{1}{c|}{Ensemble} & \multicolumn{1}{c|}{Individual} & \multicolumn{1}{c|}{Ensemble} \\ \hline
0                                              & \multicolumn{1}{l|}{0.63163}                            &                                                       & \multicolumn{1}{l|}{0.63163}                            &                                                       & \multicolumn{1}{l|}{0.63163}                                  &                                                       & \multicolumn{1}{l|}{0.63163}                            &                                                       & \multicolumn{1}{l|}{0.63163}                            &                                                       \\ \hline
1                                              & \multicolumn{1}{l|}{0.59024}                            & 0.6494                                                & \multicolumn{1}{l|}{0.61461}                            & 0.64885                                               & \multicolumn{1}{l|}{0.61452}                                  & 0.65071                                               & \multicolumn{1}{l|}{0.70889}                            & \textbf{0.70904}              & \multicolumn{1}{l|}{0.58429}                            & \textbf{0.63813}              \\ \hline
2                                              & \multicolumn{1}{l|}{0.5779}                             & 0.66526                                               & \multicolumn{1}{l|}{0.61024}                            & 0.642                                                 & \multicolumn{1}{l|}{0.62124}                                  & 0.65261                                               & \multicolumn{1}{l|}{0.60639}                            & 0.69421                                               & \multicolumn{1}{l|}{0.60553}                            & 0.62765                                               \\ \hline
3                                              & \multicolumn{1}{l|}{0.56244}                            & 0.65684                                               & \multicolumn{1}{l|}{0.5987}                             & 0.65274                                               & \multicolumn{1}{l|}{0.58315}                                  & 0.64608                                               & \multicolumn{1}{l|}{0.59152}                            & 0.68818                                               & \multicolumn{1}{l|}{0.56906}                            & 0.60846                                               \\ \hline
4                                              & \multicolumn{1}{l|}{0.58737}                            & 0.67351                                               & \multicolumn{1}{l|}{0.57877}                            & 0.66111                                               & \multicolumn{1}{l|}{0.58441}                                  & 0.65452                                               & \multicolumn{1}{l|}{0.60011}                            & 0.68156                                               & \multicolumn{1}{l|}{0.63707}                            & 0.63133                                               \\ \hline
5                                              & \multicolumn{1}{l|}{0.60204}                            & \textbf{0.67694}              & \multicolumn{1}{l|}{0.60059}                            & \textbf{0.67656}              & \multicolumn{1}{l|}{\textbf{0.68675}} & 0.65737                                               & \multicolumn{1}{l|}{0.56188}                            & 0.69705                                               & \multicolumn{1}{l|}{0.63219}                            & 0.63531                                               \\ \hline
\end{tabular}
\end{table*}

\noindent 
{\bf Constant columns degrade DNN but not XGBoost performance.} Removing constant columns/features improves the performance of TabNet. Table \ref{tab:constB} indicates that adding a constant feature column degrades the performance of TabNet significantly, while XGBoost is unaffected, as one might expect. Similar observations were made recently and discussed in \cite{shwartz2022tabular,grinsztajn2022tree,Gorishny2021revisiting}. Our industry datasets have constant columns, and removing them improves the teacher model and the best ensembles. Tables \ref{tab:cleaningB}, \ref{tab:cleaningR} and \ref{tab:cleaningV} summarize combined performance
of denoising and removing constant columns.

\begin{table*}[!ht]
\vspace{5mm}
\centering\caption{Removing and adding constant columns impacts TabNet results on Dataset B}
\label{tab:constB}
\begin{tabular}{|c|ll|ll|ll|}
\hline
    & \multicolumn{2}{l|}{Original}                                              & \multicolumn{2}{l|}{Add 5 all-one features}                                & \multicolumn{2}{l|}{Remove const features}                                 \\ \hline
Gen & \multicolumn{1}{l|}{Individual} & Ensemble         & \multicolumn{1}{l|}{Individual} & Ensemble         & \multicolumn{1}{l|}{Individual} & Ensemble         \\ \hline
0                  & \multicolumn{1}{l|}{0.56688}                            &                                          & \multicolumn{1}{l|}{0.56318}                            &                                          & \multicolumn{1}{l|}{0.6255}                             &                                          \\ \hline
1                  & \multicolumn{1}{l|}{0.60085}                            & 0.6386                                   & \multicolumn{1}{l|}{0.56646}                            & 0.57848                                  & \multicolumn{1}{l|}{0.63731}                            & 0.64534                                  \\ \hline
2                  & \multicolumn{1}{l|}{0.67693}                            & 0.70935                                  & \multicolumn{1}{l|}{0.54981}                            & 0.57872                                  & \multicolumn{1}{l|}{0.56475}                            & 0.65157                                  \\ \hline
3                  & \multicolumn{1}{l|}{0.63902}                            & 0.72011                                  & \multicolumn{1}{l|}{0.5343}                             & 0.57885                                  & \multicolumn{1}{l|}{0.56526}                            & 0.65985                                  \\ \hline
4                  & \multicolumn{1}{l|}{0.52586}                            & \textbf{0.74692} & \multicolumn{1}{l|}{0.56674}                            & 0.57992                                  & \multicolumn{1}{l|}{0.59496}                            & 0.70593                                  \\ \hline
5                  & \multicolumn{1}{l|}{0.58079}                            & 0.7463                                   & \multicolumn{1}{l|}{0.51794}                            & 0.57889                                  & \multicolumn{1}{l|}{0.64536}                            & 0.73507                                  \\ \hline
6                  & \multicolumn{1}{l|}{0.55747}                            & 0.74603                                  & \multicolumn{1}{l|}{0.55657}                            & 0.58351                                  & \multicolumn{1}{l|}{0.69509}                            & \textbf{0.76678} \\ \hline
7                  & \multicolumn{1}{l|}{0.57857}                            & 0.74603                                  & \multicolumn{1}{l|}{0.55601}                            & \textbf{0.58593} & \multicolumn{1}{l|}{0.59661}                            & 0.73137                                  \\ \hline
\end{tabular}
\end{table*}

\noindent
{\bf Order-preserving feature transforms have little effect on TabNet due to its built-in batch normalization.} Unlike models based on decision trees, DNNs tend to be sensitive to the scale
of input data. Therefore, we evaluated the impact of common transforms such as {\tt Box-Cox}, {\tt Standardize}, and {\tt Quantile} on all features. Results in Table \ref{tab:transforms} indicate that TabNet's performance is unaffected by
the {\tt Box Cox} and {\tt Quantile} transforms, while the performance dips after the {\tt Standardize} transform.
This appears related to the use of batch normalization in TabNet, the paper explains “TabNet inputs raw tabular data without any preprocessing and is trained using gradient descent-based optimization, enabling flexible integration into end-to-end learning.” Therefore, we do not apply feature transforms in subsequent experiments, but admit that individual features in some datasets could be amenable to such transforms. XGBoost is insensitive to order-preserving feature transforms, as one might expect for a model based on decision trees.
 
\begin{table*}[!ht]
\vspace{5mm}
 \centering\caption{Impact of feature transforms the quality of TabNet distillation, Dataset B}
 \label{tab:transforms}
\begin{tabular}{|c|ll|ll|ll|ll|}
\hline
 
                          & \multicolumn{2}{l|}{Original}                                              & \multicolumn{2}{l|}{w/ {\tt Standardize}}                                           & \multicolumn{2}{l|}{w/ {\tt Quantile}}                                              & \multicolumn{2}{l|}{w/ {\tt Box Cox}}                                               \\ \hline
 
Gen                       & \multicolumn{1}{l|}{Individual} & Ensemble                                 & \multicolumn{1}{l|}{Individual} & Ensemble                                 & \multicolumn{1}{l|}{Individual} & Ensemble                                 & \multicolumn{1}{l|}{Individual} & Ensemble                                 \\ \hline
0 & \multicolumn{1}{l|}{0.56688}                            &                                          & \multicolumn{1}{l|}{0.61268}                            &                                          & \multicolumn{1}{l|}{0.56688}                            &                                          & \multicolumn{1}{l|}{0.56688}                            &                                          \\ \hline
1 & \multicolumn{1}{l|}{0.60085}                            & 0.6386                                   & \multicolumn{1}{l|}{0.61748}                            & 0.63207                                  & \multicolumn{1}{l|}{0.60085}                            & 0.6386                                   & \multicolumn{1}{l|}{0.60085}                            & 0.6386                                   \\ \hline
2 & \multicolumn{1}{l|}{0.67693}                            & 0.70935                                  & \multicolumn{1}{l|}{0.57437}                            & 0.64157                                  & \multicolumn{1}{l|}{0.67693}                            & 0.70935                                  & \multicolumn{1}{l|}{0.67693}                            & 0.70935                                  \\ \hline
3 & \multicolumn{1}{l|}{0.63902}                            & 0.72011                                  & \multicolumn{1}{l|}{0.62771}                            & 0.6668                                   & \multicolumn{1}{l|}{0.63902}                            & 0.72011                                  & \multicolumn{1}{l|}{0.63902}                            & 0.72011                                  \\ \hline
4 & \multicolumn{1}{l|}{0.52586}                            & \textbf{0.74692} & \multicolumn{1}{l|}{0.65891}                            & \textbf{0.69058} & \multicolumn{1}{l|}{0.52586}                            & \textbf{0.74692} & \multicolumn{1}{l|}{0.52586}                            & \textbf{0.74692} \\ \hline
5 & \multicolumn{1}{l|}{0.58079}                            & 0.7463                                   & \multicolumn{1}{l|}{0.60042}                            & 0.66963                                  & \multicolumn{1}{l|}{0.58079}                            & 0.7463                                   & \multicolumn{1}{l|}{0.58079}                            & 0.7463                                   \\ \hline
\end{tabular}
\vspace{5mm}
\end{table*}

 \eat{
 \noindent
 {\bf Training TabNet with more epochs.} Using $\geq$100 epochs improves results for TabNet. TabNet with all discussed improvements still underperforms XGBoost on three datasets shown, but the difference is much reduced compared to our initial comparisons.
 }

\begin{table}[!ht]
\centering\caption{Removing constants and large $\Delta$s,
Dataset B (100K sampled rows) leads to small but consistent improvements
}
\label{tab:cleaningB}
\begin{tabular}{|l |l |l |l |l|}
\hline
  & TabNet  &  improved
  & XGBoost  & improved \\ 
  \hline
  & 0.68346                  &                                                                 & 0.66495                                                         &                                            \\ \hline
0 & 0.67571                  & 0.6899                                                          & 0.68899                                                         & 0.68422                                    \\ \hline
1 & 0.65953                  & 0.69008                                                         & 0.70212                                                         & 0.69445                                    \\ \hline
2 & 0.59195                  & 0.68981                                                         & 0.70517                                                         & 0.69951                                    \\ \hline
3 & 0.62141                  & { \textbf{0.69022}} & 0.70411                                                         & 0.70213                                    \\ \hline
4 & 0.57723                  & 0.68757                                                         & 0.70599                                                         & 0.70388                                    \\ \hline
5 & 0.57172                  & 0.68869                                                         & {0.70799}                                  & 0.70529                                    \\ \hline
6 & 0.57622                  & 0.68623                                                         & {\textbf{0.70833}} & 0.70633                                    \\ \hline
\end{tabular}
\end{table}
 

\subsection{DNNs vs. XGBoost on small data sets}

\begin{table}
\centering\caption{DNNs vs. XGBoost on Dataset B (10K sampled rows)}\label{tab:10K samples}
\label{tab:VS 10K}
\begin{tabular}{|c|ll|ll|}
\hline
 
\multicolumn{1}{|l|}{}           & \multicolumn{2}{l|}{TabNet}                                                & \multicolumn{2}{l|}{XGBoost}                                               \\ \hline
 
\multicolumn{1}{|l|}{Gen} & \multicolumn{1}{l|}{Individual} & Ensemble                                 & \multicolumn{1}{l|}{Individual} & Ensemble                                 \\ \hline
0                                                        & \multicolumn{1}{l|}{0.68011}                            &                                          & \multicolumn{1}{l|}{0.66868}                            &                                          \\ \hline
1                                                        & \multicolumn{1}{l|}{0.69234}                            & 0.71789                                  & \multicolumn{1}{l|}{0.68036}                            & 0.68254                                  \\ \hline
2                                                        & \multicolumn{1}{l|}{0.71806}                            & 0.75358                                  & \multicolumn{1}{l|}{0.67826}                            & \textbf{0.69169} \\ \hline
3                                                        & \multicolumn{1}{l|}{0.71872}                            & 0.76891                                  & \multicolumn{1}{l|}{0.68020}                            & 0.69137                                  \\ \hline
4                                                        & \multicolumn{1}{l|}{0.73702}                            & 0.78404                                  & \multicolumn{1}{l|}{0.65533}                            & 0.68468                                  \\ \hline
5                                                        & \multicolumn{1}{l|}{0.70074}                            & \textbf{0.79210} & \multicolumn{1}{l|}{0.68001}                            & { 0.68335}           \\ \hline
\end{tabular}
\end{table}

We found that TabNet significantly outperforms XGBoost when we train the models on smaller sample sizes. Moreover, TabNet gains a much greater boost from ensembling, as seen in Table \ref{tab:VS 10K} for 10K random samples from \texttt{Dataset B}. Yet, XGBoost generally outperforms TabNet-based models on large data sets, at least without additional optimizations for TabNet (Tables \ref{tab:optB} and \ref{tab:optV}).

To explain this phenomenon, we note the stochasticity of DNN models. Compared to XGBoost, DNNs are initialized with random weights and trained by stochastic gradient descent. This training process explores different directions in the parameter space more easily, and hence may find better model parameters when the data are limited. To verify this hypothesis, we calculate the Pearson's correlation coefficient between generations (of knowledge distillation) for both XGBoost and TabNet. The results are strikingly different. For TabNet, the correlations are $<$0.36 even for consecutive generations, while for XGBoost, the correlations are all $>$0.78. Hence, DNNs produce more diverse models across generations.
Another observation is that XGBoost tends to overfit on small datasets. In some of our experiments, XGBoost exhibited all-1 ROC AUC scores on the original training dataset across all generations, while TabNet's ROC scores range between 0.61 to 0.74, similar to the ones on the test dataset.

\subsection{Self-distillation from an ensemble}
Self-distillation trains the generation $N+1$ using generation $N$ as the teacher, then ensembles all the generations. Since ensembles usually outperform single models, one can also use a full ensemble of prior generations as a teacher to distill the next generation. Our experiments in Tables \ref{tab:XGBoost distill ens} and \ref{tab:TabNet distill ens}
compare these two options, but show no serious differences on average.
When appropriate training resources are available,
one can run both methods and pick the better model one every time.

\section{Advanced Considerations}\label{sec:advanced}
In this section, we prove equivalence between classic knowledge distillation and the more practical method introduced in Section \ref{sec:motivation} (in the context of cross-entropy loss). In Section \ref{sec:practical}, we combine this method with several other practical techniques, to be evaluated experimentally in Section \ref{sec:experiments}.

\subsection{Equivalence between distillation via weighted datasets and distillation via loss functions}
We consider $k$-class classification, $\calX\subseteq \R^d$ being the feature space, $\calY=\{1,\dots,k\}$ being the label space, and $\calD=\{\x_i,y_i\}^n_{i=1},~ (\x_i,y_i)\in\calX\times\calY$ being the training set. We consider function $f(\cdot;\theta):\calX\to\R^k$ parameterized by $\theta$ and $\smx$ being the softmax function. the loss function $\calL: \R^k\times\R^k\to \R$ is cross-entropy by default:
        $$
        \calL(\q,\p) = -\sum_{j=1}^kq_j\log p_j.
        $$
When writing $\calL(y,\p)$ for some $y\in\calY$, it refers to $\calL(\one_y,\p)$, where $\one_y$ is a one-hot vector in $\R^k$ with $y$-th entry being $1$. 
Given a teacher model $g$, \emph{Knowledge Distillation} refers to a training process for a student model $f$, where both $f$ and $g$ are functions for the $k$-class classification but they may have different architectures (hide parameter-dependency here from simplicity).
Specifically, for every $i$ defining $\q_i=\sigma(g(\x_i))$ and $\p_i=\sigma(f(\x_i))$, we define the following objective function $\calL_{kd}$ for Knowledge Distillation (KD):
        \begin{align*}
         \calL_{kd}(g,f)=\sum_{i=1}^N\left[\alpha\calL(\q_i,\p_i)+(1-\alpha)\calL(y_i,\p_i)\right].
        \end{align*}
\emph{Label Smoothing} \citep{szegedy2016rethinking} refers to a regularization method replacing $\calL(y,\sigma(f(\x)))$  with $\calL(\q^\epsilon,\sigma(f(\x)))$, where $\epsilon>0$ and $q_j^\epsilon = (1-\epsilon)\mathbbm{1}\{j=y\}+\frac{\epsilon}{k}$ for every entry $j$; in other words, $\q^\epsilon=\epsilon\bu+(1-\epsilon)\one_y$ is the ground truth vector $\one_y$ mixed with a uniform distribution $\bu$. For a single data pair $(\x,y)$, the KD loss is
        \begin{align*}
            \alpha\calL(\q,\p)+(1-\alpha)\calL(y,\p),~\\ \text{where}~\q=\sigma(g(\x)), \p=\sigma(f(\x)).
        \end{align*}
When $\calL$ is the cross-entropy, then the KD loss is equal to 
        \begin{align*}
        -\sum_{j=1}^k\log p_j\left(\alpha q_j+(1-\alpha)\mathbbm{1}\{j=y\}\right)
        \end{align*}
therefore it can be seen as $\calL(\alpha \q+(1-\alpha)\one_y,\p)$; \emph{instance-specific} label smoothing with $\q_i$ for every data $i$.

\noindent
We propose a variant of knowledge distillation via a \emph{weighted dataset}. 
Specifically, fixing an $i$, we let $$\q'_i=\alpha \q_i+(1-\alpha)\one_{y_i}.$$ 
Then we define $k$ pairs $(\x_i,j)$ with weight $q'_i(j)$. There are multiple ways to implement this. The first one is by using a weighted loss function. Specifically the weighted total loss is
\begin{align*}
\sum^N_{i=1}\left[s_i\cdot\calL(1,f_1(x_i))+(1-s_i)\cdot\calL(0,f_1(x_i))\right]
\end{align*}
Note also that:
\begin{equation*}
\begin{split}
    \calL_{wl}(g,f) &=\sum^N_{i=1}\sum^k_{j=1}q'_i(j)\calL(j,\p_i)\\
                    &=-\sum^N_{i=1}\sum_{j=1}^k\log p_i(j)\left(\alpha q_i(j)+(1-\alpha)\mathbbm{1}\{j=y_i\}\right)\\
                    &=\sum^N_{i=1}\left(\alpha\calL(\q_i,\p_i)+(1-\alpha)\calL(y_i,\p_i)\right)\\
                    &=\calL_{kd}(g,f),
\end{split}
\end{equation*}
and this again recovers knowledge distillation. However, the dataset becomes $k$ times bigger, which  significantly increases training time when there are many classes. We can therefore consider another implementation of weighted datasets by sampling. Specifically, for each instance $i$, we sample a label $z_i$ from a categorical distribution with parameter $\q'_i$ supported on $\{1,\dots,k\}$. In other words, we label $\x_i$ as class $j$ with probability $q'_i(j)$. Considering the expectation of the total loss of this implementation $\calL_{ws}(g,f)$, we see that
\begin{equation*}
\begin{split}
    \E\left[\calL_{ws}(g,f)\right]&=\E\left[\sum^N_{i=1}\calL(z_i,\p_i)\right]\\
    &=\sum^N_{i=1}\sum^k_{j=1}q'_i(j)\calL(j,\p_i)\\
    &=\calL_{wl}(g,f)=\calL_{kd}(g,f).
\end{split}
\end{equation*}

Therefore, we can view $\calL_{ws}$ as an unbiased estimator of the KD loss, which has the same size as the original dataset, and can be calculated without implementation of probabilistic labels or a customized KD loss.

Moreover, when the optimization algorithm involves stochastic gradient descent, sampling across instances is also applied by the original KD loss. In this case, our method is close to the original method but differs in how sampling is performed, which also provides an unbiased stochastic gradient. To see that, suppose we pick $I$ uniformly at random from $\{1,\dots,N\}$. Then the expectation of gradient is
\begin{equation*}
\begin{split}
    \E\left[\nabla\calL(z_I,\p_I)\right]&=\E\left[\frac{1}{N}\sum^N_{i=1}\nabla\calL(z_i,\p_i)\right]\\
    &=\frac{1}{N}\nabla\calL_{kd}(g,f).
\end{split}
\end{equation*}

\subsection{Practical techniques}
\label{sec:practical}

Based on the above equivalence, we use distillation based on weighted datasets. We perform straightforward yet impactful hyperparameter optimization, and also also optimize ensemble weights.

\noindent 
{\bf Refined distillation on input data.}
The knowledge-distillation literature uses a teacher model $f_0$ to train a student model $f_1$ with the following loss function
\begin{equation*}
    \beta\left(\sum^N_{i=1}H(f_0(x_i),f_1(x_i))\right)+(1-\beta)\left(\sum^N_{i=1}H(y_i,f_1(x_i))\right)
\end{equation*}

where $H(a,b) = - a \log b -(1-a) \log (1-b)$ is the cross entropy function. $\beta$ is a hyperparameter. When $\beta = 0$, it corresponds to the normal binary classification loss function (without any knowledge from $f_0$). In this project, we trained $f_1$ on a weighted dataset, where each $(x_i,0)$ has a weight $1-f_0(x_i)$ and each $(x_i,1)$ has a weight $f_0(x_0)$. This is similar to the case $\beta = 1$ for the loss function above. Although this forces the student model to learn and match the teacher model’s prediction on the data, it prevents the student from seeing the ground truth. We found that student models in later generations tend to perform worse. We explain that by the loss of ground-truth information during the sequential knowledge-distillation process.

Consequently, similar to the loss function, we mix the ground truth information into our weighted input. Specifically, we define positive and negative weights for each pair of data points ($(x_i,1)$ and $(x_i,0)$) as follows
\begin{align*}
    \beta f_0(x_i)+(1-\beta)y_i~\text{and}~\beta(1-f_0(x_i))+(1-\beta)(1-y_i)
\end{align*}
respectively. Likewise, $\beta = 0$ corresponds to the original dataset, and our original distillation approach corresponds to $\beta = 1$. Empirically, we found that setting $\beta = 0.7$ effectively prevents quality loss in subsequent models.

\noindent 
{\bf Tuning TabNet's hyperparameters.}
    Just like we used XGBoost ``out of the box,'' we used default hyperparameters in TabNet. The deteriorating performance of TabNet on larger datasets in our earlier experiments might be explained by the need to scale the model size and 
    architecture parameters with the size of the dataset. Therefore, we optimize
    three major parameters:
\begin{itemize}
    \item $n_a$ --- the attention layer dimension (8 by default).
    \item $n_d$ --- the prediction layer dimension (8 by default).
    \item $n_{steps}$ --- the number of decision steps in the sequential encoding process (3 by default).
\end{itemize}
Results for \texttt{Dataset V} are shown in Table \ref{tab:TabNet tuning V}.

\begin{table*}[!ht]
\centering\caption{Hyperparameter tuning for TabNet on Dataset V}
\label{tab:TabNet tuning V}
\begin{tabular}{|l |l |l |l |l |l |l |l |}
\hline
$n_d$         & 4       & 5       & 7       & 8       & 9       & 12      & 16                              \\ \hline
$n_{steps}$     & 3       & 3       & 3       & 3       & 4       & 7       & 6                               \\ \hline
\# of params & 61560   & 73370   & 97422   & 109664  & 140970  & 259080  & 311712                          \\ \hline
0            & 0.91009 & 0.91396 & 0.91959 & 0.91267 & 0.91346 & 0.90885 & 0.91816                         \\ \hline
1            & 0.91837 & 0.9217  & 0.92674 & 0.92167 & 0.91932 & 0.91935 & 0.92674                         \\ \hline
2            & 0.92251 & 0.92362 & 0.92803 & 0.92427 & 0.92297 & 0.92169 & 0.92154                         \\ \hline
3            & 0.92083 & 0.92053 & 0.92209 & 0.92181 & 0.92414 & 0.92273 & 0.92095                         \\ \hline
4            & 0.91714 & 0.92001 & 0.91858 & 0.91949 & 0.92329 & 0.92095 & 0.91837                         \\ \hline
5            & 0.91995 & 0.91673 & 0.917   & 0.91821 & 0.91976 & 0.9214  & 0.91701                         \\ \hline
Ensemble 0-5 & 0.93162 & 0.93056 & 0.93579 & 0.93362 & 0.93058 & 0.93053 & 0.93658 \\ \hline
\end{tabular}
\end{table*}

\noindent 
{\bf Optimizing ensemble weights.}
As our experimental results indicate, ensembles of teacher and student models usually outperform every individual model. Previously we obtained an ensemble by simply averaging the predictions of models. To improve upon that, we assign a weight to each model and consider weighted averages of the predictions, where the weights are optimized. The result of such optimization should not be worse than any individual model and the average ensemble we used before (but overfitting is possible with insufficient data). Optimizing ensemble weights may look like a linear problem at the first sight, but the overall classifier performance depends on ensemble weights in nonlinear ways \cite{dong2020survey}. Interactions between the (highly-correlated) models in the ensemble complicate this dependence and make optimization difficult. Ensembling models of different types tends to be more impactful than homogeneous ensembling in practice, but makes weight optimization even more challenging. 
To optimize the weights, we use the Differential Evolution method, following the discussion in \cite{brownlee2020wavgensemble}. The choice of optimization method is not essential. Tables \ref{tab:optB} and \ref{tab:optV} evaluate the optimization
of ensemble weights on two datasets, using the default TabNet architecture and one with optimized hyperparameters. The results are compared to those for XGBoost.
The entire strategy (including our prior optimizations) improves ROC AUC by 0.01–0.03 over single models. We observed that this optimization often produces some very small weights, so we round such small weights down to zero to simplify the ensembles and reduce possible overfitting (the remaining weights are re-optimized after models with zero weights are removed from the ensemble). 

\section{Empirical Validation and Deployment}
\label{sec:experiments}

We now assemble individual techniques developed and evaluated in previous sections into an overall workflow that 
produces a final high-quality model optimized for real-time
inference. This workflow is illustrated in Figure \ref{fig:workflow}. Additionally, we continue exploring comparisons between TabNet and XGBoost on data sets with 100K sampled rows (in the style of Section \ref{sec:motivation}).
In the results below, we observe that the new techniques introduced in Section \ref{sec:advanced} allow TabNet to catch up with XGBoost in performance and even beat it by a significant amount on some datasets.

\begin{table*}[!ht]
\centering\caption{Optimizing ensemble weights for the default and tuned TabNet architectures on Dataset B (100K sampled rows)
}
\label{tab:optB}
\begin{tabular}{|c|ll|ll|ll|}
\hline 
                   & \multicolumn{2}{l|}{TabNet tuned}                                                                       & \multicolumn{2}{l|}{TabNet default}                                                                     & \multicolumn{2}{l|}{}                                                                         \\ \cline{2-5}
 
                   & \multicolumn{2}{l|}{($n_d$=12, $n_a$=12, $n_{steps}$=6)}                                                     & \multicolumn{2}{l|}{($n_d$=8, $n_a$=8, $n_{steps}$ = 3)}                                                     & \multicolumn{2}{l|}{\multirow{-2}{*}{XGBoost default}}                                        \\ \cline{2-7} 

                   & \multicolumn{2}{l|}{Params: 28288}                                                                      & \multicolumn{2}{l|}{Params: 11455}                                                                      & \multicolumn{2}{l|}{Nodes: 6938–7544}                                                         \\ \cline{2-7} 
 
\multirow{-4}{*}{} & \multicolumn{1}{l|}{Single}           & Ensemble                                                        & \multicolumn{1}{l|}{Single}           & Ensemble                                                        & \multicolumn{1}{l|}{Single} & Ensemble                                                        \\ \hline
0                                          & \multicolumn{1}{l|}{\textbf{0.70157}} & \textbf{}                               & \multicolumn{1}{l|}{\textbf{0.68744}} & \textbf{}                               & \multicolumn{1}{l|}{0.67354}                        &                                                                 \\ \hline
1                                          & \multicolumn{1}{l|}{0.70878}                                  & 0.71064                                                         & \multicolumn{1}{l|}{0.67698}                                  & 0.69067                                                         & \multicolumn{1}{l|}{0.69761}                        & 0.69321                                                         \\ \hline
2                                          & \multicolumn{1}{l|}{0.71408}                                  & 0.71687                                                         & \multicolumn{1}{l|}{0.68061}                                  & \textbf{0.69452}                        & \multicolumn{1}{l|}{0.70562}                        & \textbf{0.70138}                        \\ \hline
3                                          & \multicolumn{1}{l|}{0.69722}                                  & 0.71523                                                         & \multicolumn{1}{l|}{0.6821}                                   & 0.69408                                                         & \multicolumn{1}{l|}{0.70228}                        & 0.70368                                                         \\ \hline
4                                          & \multicolumn{1}{l|}{0.6857}                                   & 0.71393                                                         & \multicolumn{1}{l|}{0.67939}                                  & 0.69226                                                         & \multicolumn{1}{l|}{0.70469}                        & 0.70515                                                         \\ \hline
5                                          & \multicolumn{1}{l|}{0.68283}                                  & \textbf{0.71314}                        & \multicolumn{1}{l|}{0.67189}                                  & { \textbf{0.69095}}                         & \multicolumn{1}{l|}{0.70371}                        & { \textbf{0.70562}} \\ \hline
Opt                                  & \multicolumn{1}{l|}{}                                         & { \textbf{0.72662}} & \multicolumn{1}{l|}{}                                         & { \textbf{0.69521}} & \multicolumn{1}{l|}{}                               &                                                                 \\ \hline
\end{tabular}
\end{table*}

\begin{table*}[!ht]
\centering\caption{Optimizing ensemble weights for the default and tuned TabNet architectures on Dataset V}
\label{tab:optV}
\begin{tabular}{|c|ll|ll|ll|}
\hline

                   & \multicolumn{2}{l|}{TabNet tuned}                                                                      & \multicolumn{2}{l|}{TabNet default}                                                                    & \multicolumn{2}{l|}{}                                  \\ \cline{2-5}

\multirow{-2}{*}{} & \multicolumn{2}{l|}{($n_d$=16, $n_a$=16, $n_{steps}$=6)}                                                    & \multicolumn{2}{l|}{($n_d$=8, $n_a$=8, $n_{steps}$ = 3)}                                                    & \multicolumn{2}{l|}{\multirow{-2}{*}{XGBoost default}} \\ \hline
 
                                           & \multicolumn{2}{l|}{Params: 311712}                                                                    & \multicolumn{2}{l|}{Params: 109664}                                                                    & \multicolumn{2}{l|}{Nodes: 9276-10488}                 \\ \hline
                                           & \multicolumn{1}{l|}{\textbf{Single}} & \textbf{Ensemble}                       & \multicolumn{1}{l|}{\textbf{Single}} & \textbf{Ensemble}                       & \multicolumn{1}{l|}{Single}      & Ensemble                                    \\ \hline
0                                          & \multicolumn{1}{l|}{0.91816}                                 &                                                                 & \multicolumn{1}{l|}{0.91267}                                 &                                                                 & \multicolumn{1}{l|}{0.94064}     &                                             \\ \hline
1                                          & \multicolumn{1}{l|}{0.92674}                                 & 0.9276                                                          & \multicolumn{1}{l|}{0.92167}                                 & \textbf{0.9235}                         & \multicolumn{1}{l|}{0.93624}     & \textbf{0.94123}    \\ \hline
2                                          & \multicolumn{1}{l|}{0.92154}                                 & 0.93151                                                         & \multicolumn{1}{l|}{0.92427}                                 & 0.92797                                                         & \multicolumn{1}{l|}{0.93177}     & 0.93964                                     \\ \hline
3                                          & \multicolumn{1}{l|}{0.92095}                                 & 0.93406                                                         & \multicolumn{1}{l|}{0.92181}                                 & 0.9303                                                          & \multicolumn{1}{l|}{0.92886}     & 0.93795                                     \\ \hline
4                                          & \multicolumn{1}{l|}{0.91837}                                 & \textbf{0.9351}                         & \multicolumn{1}{l|}{0.91949}                                 & { \textbf{0.93209}}                         & \multicolumn{1}{l|}{0.92699}     & 0.93648                                     \\ \hline
5                                          & \multicolumn{1}{l|}{0.91701}                                 & 0.93658                                                         & \multicolumn{1}{l|}{0.91821}                                 & 0.93362                                                         & \multicolumn{1}{l|}{0.92568}     & 0.93524                                     \\ \hline
Opt                                  & \multicolumn{1}{l|}{}                                        & { \textbf{0.93692}} & \multicolumn{1}{l|}{}                                        & { \textbf{0.93391}} & \multicolumn{1}{l|}{}            &                                             \\ \hline
\end{tabular}
\vspace{5mm}
\end{table*}

\subsection{Combining XGBoost and TabNet}
Hoping for further improvement, we combine TabNet and XGBoost.
Similar to the optimizing ensemble weights technique in the previous section, here we ensemble generations from both XGBoost and TabNet.
We obtain a score better than every single ensemble we observed --- not a surprising result, but significant nevertheless.
Upon careful inspection, we observe that only the first few XGBoost generations are used. This reconfirms a more general trend we've discussed --- DNNs gain more from knowledge distillation than XGBoost models do.

\subsection{Optimizing XGBoost hyperparameters}

Now we also optimize XGBoost, since we used default hyperparameters so far. $n_{estimators}$ is the number of boosting rounds and thus the number of trees built. In Table \ref{tab:xgb_n_est}, we compare the default XGBoost (100 trees) to XGBoost models with 200, 400, and 1000 trees.  As $n_{estimators}$ grows, model performance continues improving, seemingly without saturation. With 1000 trees, it even beats the ensemble of the original XGBoost and TabNet we showed above, confirming that XGBoost is very competitive for tabular data. Tuning additional hyperparameters may further improve results, but risks
diminishing returns.

\begin{table}[!ht]
\centering
\caption{Optimizing XGBoost on Dataset V}
\label{tab:xgb_n_est}
\begin{tabular}{|l|ll|ll|}
\hline
 
\# of Trees   & \multicolumn{2}{l|}{100}                                                         & \multicolumn{2}{l|}{400}                                                         \\ \hline

Avg Nodes & \multicolumn{1}{l|}{10000}            & 18000                                    & \multicolumn{1}{l|}{34500}            & 83000                                    \\ \hline
0             & \multicolumn{1}{l|}{\textbf{0.94064}} & \textbf{0.94454} & \multicolumn{1}{l|}{\textbf{0.94848}} & \textbf{0.9528}  \\ \hline
1             & \multicolumn{1}{l|}{0.93624}                                  & 0.94277                                  & \multicolumn{1}{l|}{0.94767}                                  & 0.95268                                  \\ \hline
2             & \multicolumn{1}{l|}{0.93177}                                  & 0.94008                                  & \multicolumn{1}{l|}{0.94798}                                  & \textbf{0.95113} \\ \hline
3             & \multicolumn{1}{l|}{0.92886}                                  & 0.93771                                  & \multicolumn{1}{l|}{0.94556}                                  & 0.95179                                  \\ \hline
4             & \multicolumn{1}{l|}{0.92699}                                  & 0.93629                                  & \multicolumn{1}{l|}{0.94543}                                  & 0.95205                                  \\ \hline
5             & \multicolumn{1}{l|}{0.92568}                                  & \textbf{0.93333} & \multicolumn{1}{l|}{0.94457}                                  & { \textbf{0.95101}}  \\ \hline
\end{tabular}
\end{table}

\subsection{Deployment considerations}
As seen in Section \ref{sec:experiments}, with multiple advanced training techniques, TabNet ensembles become competitive with XGBoost, and combining such models further improves resuts. As these results are validated on data from the industry \Looper platform, we explore application deployment with an eye on any obstacles. To this end, we implemented all proposed techniques within the training workflow in \Looper. However, the infrastructure available to us does not currently support
highly-optimized real-time inference for TabNet and ensembles of multiple models.
Rather than implement such support (a considerable ML infrastructure project), 
we propose a shortcut --- training a comparable model of supported type.
Here we again rely on knowledge distillation: we distill the best available ensemble model into a simpler model supported for high-performance real-time inference via our distillation via a weighted dataset. This can be accomplished by creating a weighted dataset table and launching an appropriate API-driven model flow ---- a routine task, much simpler than implementing the entire self-distillation procedure from scratch. More importantly, the resulting model exhibits practically the same performance and preserves the consistent improvements we have observed for more sophisticated (best) models across the \Looper use cases we worked with. Results in Table \ref{tab:final} show compelling ROC AUC improvements for all use cases. Gains range from 0.07\% for {\tt Dataset R} to 7.14\% for {\tt Dataset C}. Large gains are mostly associated with initial models that don't exhibit strong performance, whereas improving strong models is difficult (no surprises here).

\begin{table}[bh]
\vspace{-4mm}
\caption{Shapes of industry use cases. Letter labels are derived from full names of company-internal use cases, rather than taken as consecutive letters of the alphabet (e.g., there is no Dataset D).}
\label{tab:meta data}
\vspace{-1mm}
\centering
\begin{tabular}{|l|c|cccc|}
\hline
                                         &                                  & \multicolumn{4}{c|}{\# of each type of features}                                                                                                                        \\ \cline{3-6} 
\multirow{-2}{*}{Use Case}               & \multirow{-2}{*}{\# examples} & \multicolumn{1}{c|}{bool} & \multicolumn{1}{c|}{int} & \multicolumn{1}{c|}{float} & categorical \\ \hline
Dataset A                            & \textbf{19,835}                                          & \multicolumn{1}{c|}{1}    & \multicolumn{1}{c|}{2}   & \multicolumn{1}{c|}{9}                             & 0                                   \\ \hline

Dataset B     & \textbf{1,202,700}                                       & \multicolumn{1}{c|}{0}    & \multicolumn{1}{c|}{38}  & \multicolumn{1}{c|}{23}                            & 5                                   \\ \hline

Dataset C & \textbf{2,338}                                                    & \multicolumn{1}{c|}{5}                            & \multicolumn{1}{c|}{58}                          & \multicolumn{1}{c|}{6}                             & 10                                  \\ \hline

Dataset I     & \textbf{186,854}                                         & \multicolumn{1}{c|}{4}    & \multicolumn{1}{c|}{16}  & \multicolumn{1}{c|}{612}                           & 9                                   \\ \hline

Dataset J      & \textbf{1,291,701}                                       & \multicolumn{1}{c|}{15}   & \multicolumn{1}{c|}{65}  & \multicolumn{1}{c|}{10}                            & 15                                  \\ \hline


Dataset P   & \textbf{127,850}                                         & \multicolumn{1}{c|}{50}   & \multicolumn{1}{c|}{121} & \multicolumn{1}{c|}{79}                            & 20                                  \\ \hline

Dataset R                         & \textbf{2,118,670}                                       & \multicolumn{1}{c|}{4}    & \multicolumn{1}{c|}{37}  & \multicolumn{1}{c|}{13}                            & 8                                   \\ \hline
Dataset S                        & \textbf{1,032,414}                                       & \multicolumn{1}{c|}{5}    & \multicolumn{1}{c|}{7}   & \multicolumn{1}{c|}{20}                            & 10                               
\\ \hline

Dataset V          & \textbf{103,883}                                          & \multicolumn{1}{c|}{7}    & \multicolumn{1}{c|}{66}   & \multicolumn{1}{c|}{46}                             & 5                                \\ \hline

Dataset W          & \textbf{73,100}                                          & \multicolumn{1}{c|}{2}    & \multicolumn{1}{c|}{9}   & \multicolumn{1}{c|}{2}                             & 10                                  \\ \hline

\end{tabular}
\end{table}


\begin{table}[!ht]
\caption{The ROC AUC scores for the use cases. 
}
\label{tab:final}
\centering
\begin{tabular}{|l|c |c |c |l}
\cline{1-4}
\multicolumn{1}{|c|}{Use Case}           & Ours & Baseline & Gain &  \\ \cline{1-4}
Dataset A                            & \textbf{0.89828}             & 0.89555                          & 0.27\%                       &  \\ \cline{1-4}

Dataset B     & \textbf{0.70414}             & 0.68417                          & 2.00\%                       &  \\ \cline{1-4}

Dataset C & \textbf{0.91071}             & 0.83928                          & 7.14\%                       &  \\ \cline{1-4}

Dataset I     & \textbf{0.98536}             & 0.98520                          & 0.02\%                       &  \\ \cline{1-4}

Dataset J      & \textbf{0.98320}             & 0.98258                          & 0.06\%                       &  \\ \cline{1-4}


Dataset P   & \textbf{0.63069}             & 0.59525                          & 3.54\%                       &  \\ \cline{1-4}

Dataset R                         & \textbf{0.87349}             & 0.87276                          & 0.07\%                       &  \\ \cline{1-4}
Dataset S                       & \textbf{0.70414}             & 0.68417                          & 2.00\%                       &  \\ \cline{1-4}

Dataset W          & \textbf{0.64012}             & 0.63156                          & 0.86\%                       &  \\ \cline{1-4}
\end{tabular}

\end{table}

 \begin{table}[!htb]
 \centering\caption{Removing constant columns and large $\Delta$s on Dataset V leads to small but consistent improvements.}
 \label{tab:cleaningV}
\vspace{-1mm}
\begin{tabular}{|l|l|l|l|l|}
\hline
     Gen              &  TabNet        &   improved                                &  XGBoost  &   improved                      \\ 
\hline
0                                          & 0.90733                  &                                                                 & 0.93608                                           &                                                                 \\ \hline
1                                          & 0.90965                  & 0.91146                                                         & 0.93362                                           & { \textbf{0.93784}} \\ \hline
2                                          & 0.91034                  & 0.91314                                                         & 0.93033                                           & 0.93693                                                         \\ \hline
3                                          & 0.91057                  & 0.91414                                                         & 0.92783                                           & 0.93568                                                         \\ \hline
4                                          & 0.90888                  & { \textbf{0.91439}} & 0.92562                                           & 0.93439                                                         \\ \hline
5                                          & 0.90816                  & 0.91435                                                         & 0.92363                                           & 0.93326                                                         \\ \hline
\end{tabular}
\end{table}

\begin{table}[!ht]
\centering\caption{XGBoost distilled from last vs. from ensemble on Dataset B. No substantial difference in results is observed,}
\label{tab:XGBoost distill ens}
\vspace{-1mm}
\begin{tabular}{|c|ll|ll|}
\hline 
\multicolumn{1}{|l|}{}           & \multicolumn{2}{l|}{XGBoost (dist. from last)}                           & \multicolumn{2}{l|}{XGBoost (dist. from ens.)}                       \\ \hline
\multicolumn{1}{|l|}{Gen} & \multicolumn{1}{l|}{Individual} & Ensemble                                 & \multicolumn{1}{l|}{Individual} & Ensemble                                 \\ \hline
0                                                        & \multicolumn{1}{l|}{0.89705}                            &                                          & \multicolumn{1}{l|}{0.89705}                            &                                          \\ \hline
1                                                        & \multicolumn{1}{l|}{0.89864}                            & 0.90038                                  & \multicolumn{1}{l|}{0.89864}                            & 0.90038                                  \\ \hline
2                                                        & \multicolumn{1}{l|}{0.89986}                            & 0.90108                                  & \multicolumn{1}{l|}{0.90018}                            & \textbf{0.90155} \\ \hline
3                                                        & \multicolumn{1}{l|}{0.90062}                            & 0.90164                                  & \multicolumn{1}{l|}{0.90030}                            & 0.90190                                  \\ \hline
4                                                        & \multicolumn{1}{l|}{0.90080}                            & 0.90212                                  & \multicolumn{1}{l|}{0.90019}                            & 0.90201                                  \\ \hline
5                                                        & \multicolumn{1}{l|}{0.89997}                            & \textbf{0.90218} & \multicolumn{1}{l|}{0.90057}                            & { \textbf{0.90222}}  \\ \hline
\end{tabular}
\end{table}

\begin{table}[!ht]
\centering\caption{TabNet distilled from last vs. from ensemble on Dataset B. No substantial difference in results is observed,}
\label{tab:TabNet distill ens}
\begin{tabular}{|c|ll|ll|}
\hline
\multicolumn{1}{|l|}{}           & \multicolumn{2}{l|}{TabNet (dist. from last)}                            & \multicolumn{2}{l|}{TabNet (dist. from ens.)}                        \\ \hline
\multicolumn{1}{|l|}{Gen} & \multicolumn{1}{l|}{Individual} & Ensemble                                 & \multicolumn{1}{l|}{Individual} & Ensemble                                 \\ \hline
0                                                        & \multicolumn{1}{l|}{0.88899}                            &                                          & \multicolumn{1}{l|}{0.88899}                            &                                          \\ \hline
1                                                        & \multicolumn{1}{l|}{0.89321}                            & 0.89462                                  & \multicolumn{1}{l|}{0.89321}                            & 0.89462                                  \\ \hline
2                                                        & \multicolumn{1}{l|}{0.89316}                            & 0.89535                                  & \multicolumn{1}{l|}{0.89483}                            & \textbf{0.89659} \\ \hline
3                                                        & \multicolumn{1}{l|}{0.89489}                            & 0.89682                                  & \multicolumn{1}{l|}{0.89297}                            & 0.89720                                  \\ \hline
4                                                        & \multicolumn{1}{l|}{0.89520}                            & 0.89750                                  & \multicolumn{1}{l|}{0.89411}                            & 0.89789                                  \\ \hline
5                                                        & \multicolumn{1}{l|}{0.89418}                            & \textbf{0.89791} & \multicolumn{1}{l|}{0.89360}                            & { \textbf{0.89833}}  \\ \hline
\end{tabular}
\end{table}

\begin{table*}[!ht]
\centering\caption{Removing constants and large $\Delta$s on
Dataset R (100K sampled rows) leads to improvements
}
\label{tab:cleaningR}
\vspace{-1mm}
\begin{tabular}{|l |l |l |l |l |l |l |}
\hline
                   & TabNet  &    $\Delta=0.001$                              
         & TabNet  &   $\Delta=0.01$                                       &  XGBoost &                                         \\
\hline
0                                          & 0.87384                        &                                                                 & 0.87384                       &                                                                 & 0.88393                                           &                                                                 \\ \hline
1                                          & 0.87543                        & 0.87556                                                         & 0.87534                       & 0.87562                                                         & 0.88631                                           & 0.88811                                                         \\ \hline
2                                          & 0.87563                        & 0.87623                                                         & 0.8755                        & 0.87578                                                         & 0.88653                                           & 0.88911                                                         \\ \hline
3                                          & 0.87558                        & 0.8766                                                          & 0.87556                       & 0.87602                                                         & 0.8852                                            & { \textbf{0.88922}} \\ \hline
4                                          & 0.87514                        & { \textbf{0.87663}} & 0.87571                       & 0.87621                                                         & 0.88338                                           & 0.88899                                                         \\ \hline
5                                          & 0.8746                         & 0.87658                                                         & 0.87561                       & { \textbf{0.87643}} & 0.88284                                           & 0.88872                                                         \\ \hline
\end{tabular}
\end{table*}

\section{Conclusions and Future Work}
We developed a self-distillation method, made it practical, and successfully validated it on data from an industry high-performance end-to-end real-time ML platform.
As theoretical justification for it, we proved the equivalence between our approach and the classical knowledge distillation in terms of (the expectation of) the loss function. In addition to using self-distillation
to produce powerful ensembles of DNNs and optimizing those ensembles using
Differential Evolution, we distill these ensembles into simpler models appropriate for high-performance real-time inference in 
practice.

What we have accomplished looks paradoxical. Recall that the prior {\em status quo} had DNN models such as TabNet lose to gradient boosting on tabular data \citep{grinsztajn2022tree,shwartz2022tabular,Gorishny2021revisiting}. Since knowledge distillation improves TabNet more than it improves XGBoost, twe reduce he performance gap, and TabNet ensembles start winning on small data sets because they leverage diversity and avoid overfitting. With additional enhancements, TabNet ensembles are improved further and exhibit superior performance. However, their implementation complexity obstructs their use in high-performance real-time inference in the end-to-end ML platform. Therefore, we distill these advanced models into compact XGBoost models. Thus, we have used DNNs to help XGBoost beat (single) DNNs by a larger margin on tabular data and then match the performance of powerful DNN ensembles. For large non-tabular datasets, gradient-boosting models might lack the capacity to compete with DNN architectures. However, our tabular datasets reach into respectable sizes, and we did not observe such trends. The results presented in this paper use proprietary datasets, but the
trends we report, in all likelihood, carry over to public tabular data sets used in \citep{grinsztajn2022tree,shwartz2022tabular,Gorishny2021revisiting}.

Our techniques and results involving gradient boosting on tabular data may be of particular interest in the context of resource-constrained high-performance inference on tabular data. On the other hand, our empirical results are limited to data sets without sparse features, such as any kind of ID numbers, because XGBoost does handle them well. Today, sparse features are best handled by trained latent-space embeddings within specialized DNN architectures, and such extensions can be adapted to TabNet. To this end, our data denoising and distillation techniques should carry over verbatim to produce DNN ensembles with improved performance. Gradient boosting models lack comparable facilities for sparse features, but can be combined with DNN architectures through {\em stacking}, i.e. by considering tree leaves as features for DNNs. Overall, our techniques can be viewed in a larger universe of hybrid ML models for supervised learning. In addition, recent findings on knowledge distillation in the context of Reinforcement Learning (RL) \citep{li2021neural} suggest adapting our techniques to RL applications, e.g., \citep{https://doi.org/10.48550/arxiv.2102.05612}.


\bibliographystyle{ACM-Reference-Format}
\bibliography{sample-base}

\end{document}